\documentclass[11pt]{article}
\usepackage[utf8]{inputenc}
\usepackage[T1]{fontenc}
\usepackage{fixltx2e}
\usepackage{graphicx}
\usepackage{longtable}
\usepackage{float}
\usepackage{wrapfig}
\usepackage{soul}
\usepackage{textcomp}
\usepackage{marvosym}
\usepackage[integrals]{wasysym}
\usepackage{latexsym}
\usepackage{amssymb}
\usepackage{hyperref}
\usepackage{amsmath}
\usepackage{nips12submit_e}
\usepackage{times}
\usepackage{caption}
\usepackage{subcaption}
\usepackage{stmaryrd}
\DeclareMathOperator*{\argmin}{arg\,min}

\nipsfinalcopy

\begin{document}

\title{Switched linear coding with rectified linear autoencoders}
\author{
Leif Johnson \qquad Craig Corcoran \\
Computer Science Department \\
The University of Texas at Austin \\
Austin, TX 78701 \\
\texttt{\{leif,ccor\}@cs.utexas.edu}}
\date{January 17, 2013}
\maketitle

\begin{abstract}
 Several recent results in machine learning have established formal connections
 between autoencoders---artificial neural network models that attempt to
 reproduce their inputs---and other coding models like sparse coding and
 K-means. This paper explores in depth an autoencoder model that is constructed
 using rectified linear activations on its hidden units. Our analysis builds on
 recent results to further unify the world of sparse linear coding models. We
 provide an intuitive interpretation of the behavior of these coding models and
 demonstrate this intuition using small, artificial datasets with known
 distributions.
\end{abstract}

\section{Introduction}
\label{sec-1}

Large quantities of natural data are now commonplace in the digital
world---images, videos, and sounds are all relatively cheap and easy to obtain.
In comparison, labels for these datasets (e.g., answers to questions like, ``Is
this an image of a cow or a horse?'') remain relatively expensive and difficult
to assemble. In addition, even when labeled data are available for a given task,
there are often only a few bits of information in the labels, while the
unlabeled data can easily contain millions of bits. Finally, most collections of
data from the natural world also seem to be distributed non-uniformly in the
space of all possible inputs, suggesting that there is some underlying structure
inherent in a particular dataset, independent of labels or task. Furthermore,
many unsupervised learning approaches assume that natural data even lie on a
relatively continuous low-dimensional manifold of the available space, which
provides further structure that can be captured in a model. For these reasons,
much recent work in machine learning has focused on unsupervised modeling of
large, easy to obtain datasets.

An unsupervised model attempts to create a compact representation of the
structure of a dataset. Some models are designed to learn a set of ``basis
functions'' that can be combined linearly to create the observed data. Once such
a set has been learned, this representation is often useful in other tasks like
classification or recognition (but see \cite{denil2012recklessly} for
interesting analysis). In addition to learning useful representations, many
models that have resulted from this line of work have also been shown to have
interesting theoretical ties with neuroscience and information theory (e.g.,
\cite{olshausen1996emergence}).

In this paper, we first synthesize recent results from unsupervised machine
learning, with a particular focus on neural networks and sparse coding. We then
explore in detail the coding behavior of one important model in this class:
autoencoders constructed with rectified linear hidden units. After describing
the intuitive behavior of these models, we examine their behavior on artificial
and natural datasets.
\section{Background}
\label{sec-2}

One way to take advantage of a readily available unlabeled dataset $\mathcal{O}$
is to learn features that encode the data well. Such features should then
presumably be useful for describing the data effectively, regardless of the
particular task. Formally, we wish to find a ``dictionary'' or ``codebook'' matrix
$D \in \mathbb{R}^{n \times k}$ whose columns can be combined or ``decoded''
linearly to represent elements from the dataset. There are many ways to do this:
for example, one could use random vectors or samples from $\mathcal{O}$, or
define some cost function over the space of dictionaries and optimize to find a
good one.

Once $D$ is known, then for any input ${\bf x}$, we would also like to compute
the coefficients ${\bf h} \in \mathbb{R}^k$ that give the best linear
approximation $\hat{\bf x} = D{\bf h}$ to ${\bf x}$. This problem is often
expressed as an optimization of a squared error loss \[ {\bf h} = \argmin_{\bf
u} \| D{\bf u} - {\bf x} \|^2_2 + R({\bf u}) \] where $R$ is some regularizer.
If $R=0$, for example, then this is linear regression; when $R=\|\cdot\|_1$, the
encoding problem is known as sparse coding or lasso regression
\cite{tibshirani1996regression}. Sparse coding yields state-of-the art results
on many tasks for a wide variety of dictionary learning methods
\cite{coates2011importance}, but it requires a full optimization procedure to
compute ${\bf h}$, which might require more or less computation depending on the
input. To address this complexity, Gregor and LeCun \cite{gregor2010learning}
tried approximating this coding procedure by training a neural network (which
has a bounded complexity) on a dataset labeled with its sparse codes. For the
remainder of the paper, we consider a similar problem---learning an efficient
code for a set of data---but focus on simpler autoencoder neural network models
to avoid solving the complete sparse coding optimization problem.
\subsection{Autoencoders}
\label{sec-2_1}

Autoencoders \cite{hinton1994autoencoders, vincent2008extracting,
lemme2010efficient} are a versatile family of neural network models that can be
used to learn a dictionary from a set of unlabeled data, and to compute an
encoding ${\bf h}$ for an input ${\bf x}$. This learning takes place
simultaneously, since changing the dictionary changes the optimal encodings. An
autoencoder attempts to reconstruct its input after passing activation forward
through one or more nonlinear hidden layers. The general loss function for a
one-layer autoencoder can be written as \[
 \ell(\mathcal{O}) = \frac{1}{2\left|\mathcal{O}\right|} \sum_{{\bf
 x}\in\mathcal{O}} \left\| \varsigma \left( D \sigma\left( W{\bf x} + {\bf
 b}\right) \right) - {\bf x} \right\|_2^2 + R(\mathcal{O}, D, W, {\bf b})
\] where $\varsigma(\cdot)$ and $\sigma(\cdot)$ are activation functions for the
output and hidden neurons, respectively, ${\bf b}\in\mathbb{R}^k$ is a vector of
hidden unit activation biases, and $W\in\mathbb{R}^{k \times n}$ is a matrix of
weights for the hidden units. By setting $\varsigma(z)=z$, this family of models
clearly belongs to the class of coding models, where ${\bf h} = \sigma(W{\bf
x} + {\bf b})$. This encoding scheme is easy to compute, but it also assumes
that the optimal coefficients can be represented as a linear transform of the
input, passed through some nonlinearity.

Commonly, the encoding and decoding weights are ``tied'' so that $W=D^T$, which
forces the dictionary to take on values that are useful for both tasks. Seen
another way, tied weights provide an implicit orthonormality constraint on the
dictionary \cite{le2011ica}. Tying the weights also reduces by half the number
of parameters that the model needs to tune.
\subsection{Rectified linear autoencoders}
\label{sec-2_2}

Traditionally, neural networks are constructed with sigmoid activations like
$\sigma(z) = \left(1 + e^z\right)^{-1}$. The rectified linear activation
function $[z]_+ = \max(0, z)$ has been shown to improve training and performance
in multilayer networks by addressing several shortcomings of sigmoid activation
functions \cite{nair2010rectified, glorot2011deep}. In particular, the rectified
linear activation function produces a true 0 (not just a small value) for
negative input and then increases linearly. Unlike the sigmoid, which has a
derivative that vanishes for large input, the derivative of the rectified linear
activation is either 0 for negative inputs, or 1 for positive inputs This
``switching'' behavior is inspired by the firing rate pattern seen in some
biological neurons, which are inactive for sub-threshold inputs and then
increase firing rate as input magnitude increases. The true 0 output of the
rectified linear activation function effectively deactivates neurons that are
not tuned for a specific input, isolating a smaller active model for that input
\cite{glorot2011deep}. This also has the effect of combining exponentially many
linear coding schemes into one codebook \cite{nair2010rectified}.

To learn a dictionary from data, Glorot et al. \cite{glorot2011deep} proposed
the single-layer rectified linear autoencoder with tied weights by setting
$W=D^T$, $\varsigma(z)=z$ and $\sigma(z)=[z]_+$ in the general autoencoder loss,
giving \[
 \ell(\mathcal{O}) = \frac{1}{2\left|\mathcal{O}\right|} \sum_{{\bf x} \in
 \mathcal{O}} \left\| D \left[ D^T{\bf x} + {\bf b} \right]_+ - {\bf x}
 \right\|_2^2 + R(\mathcal{O}, D, {\bf b}).
\]

Hidden units in rectified linear autoencoders behave similarly to the ``triangle
K-means'' encoding proposed by Coates et al. \cite{coates2011analysis}, which
produces a coefficient vector ${\bf h}$ for input ${\bf x}$ such that $h_i =
\left[ \; \mu - \| {\bf x} - {\bf c}_i \| \; \right]_+$, where ${\bf \mu} =
\frac{1}{k} \sum_{j=1}^k \|{\bf x} - {\bf c}_j\|$ is the mean distance to all
cluster centroids. Like a rectified linear model, triangle K-means produces true
0 activations for many (approximately half) of the clusters, while coding the
rest linearly. An equivalent \cite{denil2012recklessly} variation is the ``soft
thresholding'' scheme \cite{coates2011importance}, where $h_i = \left[ {\bf
d}_i^T{\bf x} - \lambda \right]_+$ for some parameter $\lambda$. This coding
scheme is equivalent to a rectified linear autoencoder with a shared bias
$\lambda$ on all hidden units. Rectified linear autoencoders generalize all of
the models in this ``switched linear'' class and provide a unified form for their
loss functions.
\subsection{Whitening}
\label{sec-2_3}

The switched linear family of models uses linear codes for some regions of input
space and produces zeros in others. These models perform best when applied to
whitened datasets. In fact, Le et al. \cite{le2011ica} showed that sparse
autoencoders, sparse coding, and ICA all compute the same family of loss
functions, but only if the input data are approximately white. Empirically,
Coates et al. \cite{coates2011analysis, coates2011importance} observed that for
several different single-layer coding models, switched-linear coding schemes
perform best when combined with pre-whitened data.

Whitening seems to be important in biological systems as well: Although the
human eye does not receive white input, neurons in retina and LGN appear to
whiten local patches of natural images \cite{simoncelli2001natural,
dan1996efficient} before further processing in the visual cortex.
\section{Switched linear codes}
\label{sec-3}

\begin{figure*}[tb]
\centering
\begin{subfigure}{0.45\textwidth}
 \centering
 \includegraphics[width=\textwidth,clip=true,trim=40mm 105mm 40mm 90mm]{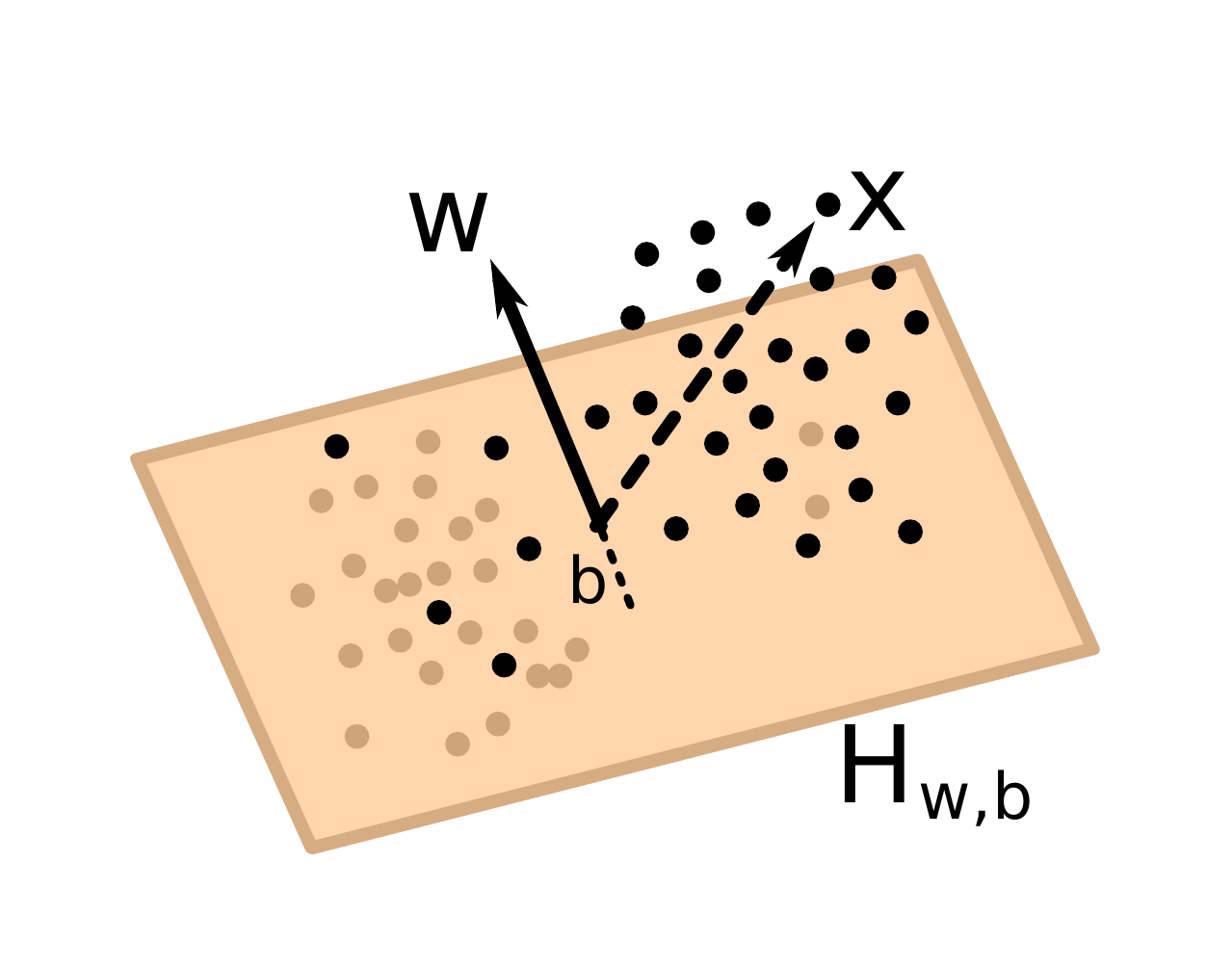}
 \caption{Classifier view.}
\end{subfigure}
\begin{subfigure}{0.45\textwidth}
 \centering
 \includegraphics[width=\textwidth,clip=true,trim=40mm 105mm 40mm 90mm]{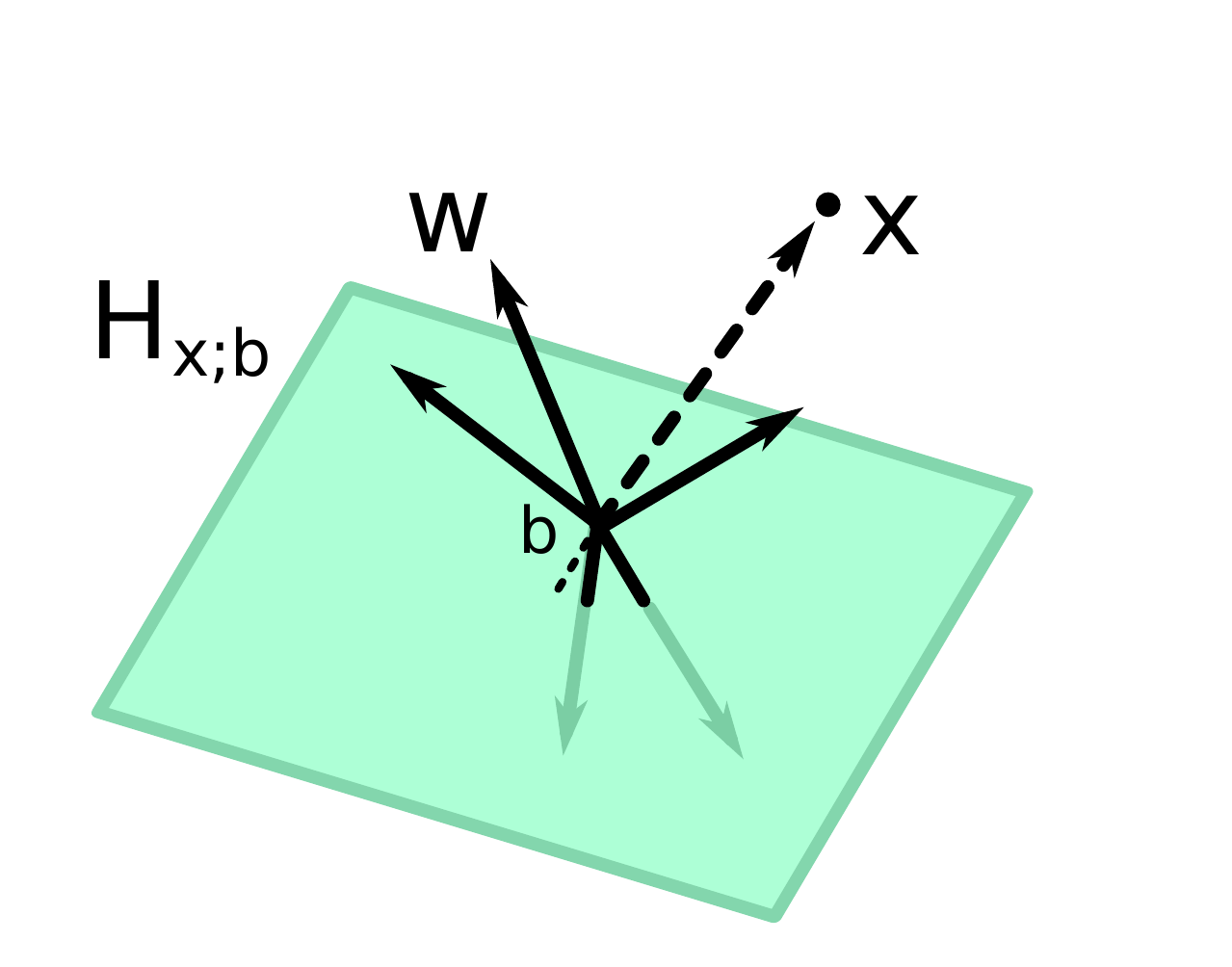}
 \caption{Encoder view.}
\end{subfigure}
\caption{Visualizing the half-spaces associated with data point ${\bf x}$ and
 feature vector ${\bf w}$. In a linear classifier (left), weight vectors define
 hyperplanes that split data points into two groups. In a linear encoder
 (right), a data point ${\bf x}$ defines a hyperplane which is shifted in the
 direction opposite that data point through a distance $b$ associated with
 feature ${\bf w}$; the hidden unit corresponding to ${\bf w}$ is coded by the
 distance from ${\bf w}$ to the shifted hyperplane.}
\label{fig:halfspace}
\end{figure*}

Many researchers (e.g., \cite{hahnloser2003permitted, nair2010rectified,
glorot2011deep}) have described how the rectified linear activation function
separates its inputs into two natural groups of outputs: those with 0 values,
and those with positive values. We belabor this observation here to make a
further point about the behavior of rectified linear autoencoders below.

Consider a data point ${\bf x}$ and each encoding feature (row) ${\bf w}_j$ of
$D^T$ geometrically as vectors in $\mathbb{R}^n$, shown in Figure
\ref{fig:halfspace}. In the context of a linear classifier, the weight vector
${\bf w}$ is often described as defining the normal to a hyperplane $H_{\bf w}$
that divides the dataset into two subsets. A linear encoder, on the other hand,
is attempting to determine which hidden units describe its input, so in this
context it is the vector ${\bf x}$ that defines the normal to a hyperplane
$H_{\bf x}$ passing through the origin. Each feature ${\bf w}_j$ in the
autoencoder corresponds to a bias term $b_j$; this bias can be seen as
translating $H_{\bf x}$ in the direction away from ${\bf x}$ through a distance
of $b_j$. After this translation, if ${\bf w}_j$ lies on the opposite side of
$H_{{\bf x};{\bf w}_j,b_j}$ from ${\bf x}$, then the hidden unit $h_j$ will be
set to 0 by a rectified activation function.\footnote{Analytically, $z_j = {\bf
w}_j {\bf x} + b_j < 0$, so $h_j = \max(0, z_j) = 0$.} In other words,
rectification selectively deactivates features ${\bf w}_j$ in the loss for ${\bf
x}$ whenever feature ${\bf w}_j$ points sufficiently far away from ${\bf
x}$.\footnote{Where "sufficiently" is determined by the bias for that feature.}
After hidden unit $h_j$ is deactivated, decoding entry ${\bf w}_j$ no longer has
any effect on the output. We can thus define \[ \psi_{\bf x} = \left\{ j:{\bf
w}_j{\bf x} + b_j > 0 \right\} \] as the set of features that have nonzero
activations in the hidden layer of the autoencoder for input ${\bf x}$. Then \[
{\bf h}_\psi = \left[ h_{\psi_1}, \dots, h_{\psi_r} \right]^T \] is the vector
of hidden activations for just the features in $\psi$, $D_\psi$ is the matrix of
corresponding columns from $D$, and ${\bf b}_\psi$ are the corresponding bias
terms. This gives a new loss function \[
 \ell_\psi(\mathcal{O}) = \frac{1}{2\left|\mathcal{O}\right|} \sum_{{\bf
 x}\in\mathcal{O}} \left\| D_\psi \left( D^T_\psi{\bf x} + {\bf b}_\psi\right) -
 {\bf x} \right\|_2^2 + R(\mathcal{O}, D, {\bf b})
\] that is computed only over the active features $\psi_{\bf x}$ for each input
${\bf x}$.

With ${\bf b} = {\bf 0}$ and a sparse regularizer $R = \|{\bf h}\|_1$, the loss
$\ell_\psi$ reduces to that proposed by Le et al. \cite{le2011ica} for ICA with
a reconstruction cost. For whitened input, then, the rectified linear
autoencoder behaves not only like a mixture of exponentially many linear models
\cite{nair2010rectified}, but rather like a mixture of exponentially many ICA
models---one ICA model is selected for each input ${\bf x}$, but all features
are selected from, and must coexist in, one large feature set. Importantly, this
whitening requirement only holds for the region of data delimited by the set of
active features, potentially simplifying the whitening transform.

\begin{figure*}[tb]
\centering
\begin{subfigure}{0.5\textwidth}
 \centering
 \includegraphics[width=\textwidth,clip=true,trim=20mm 10mm 20mm 10mm]{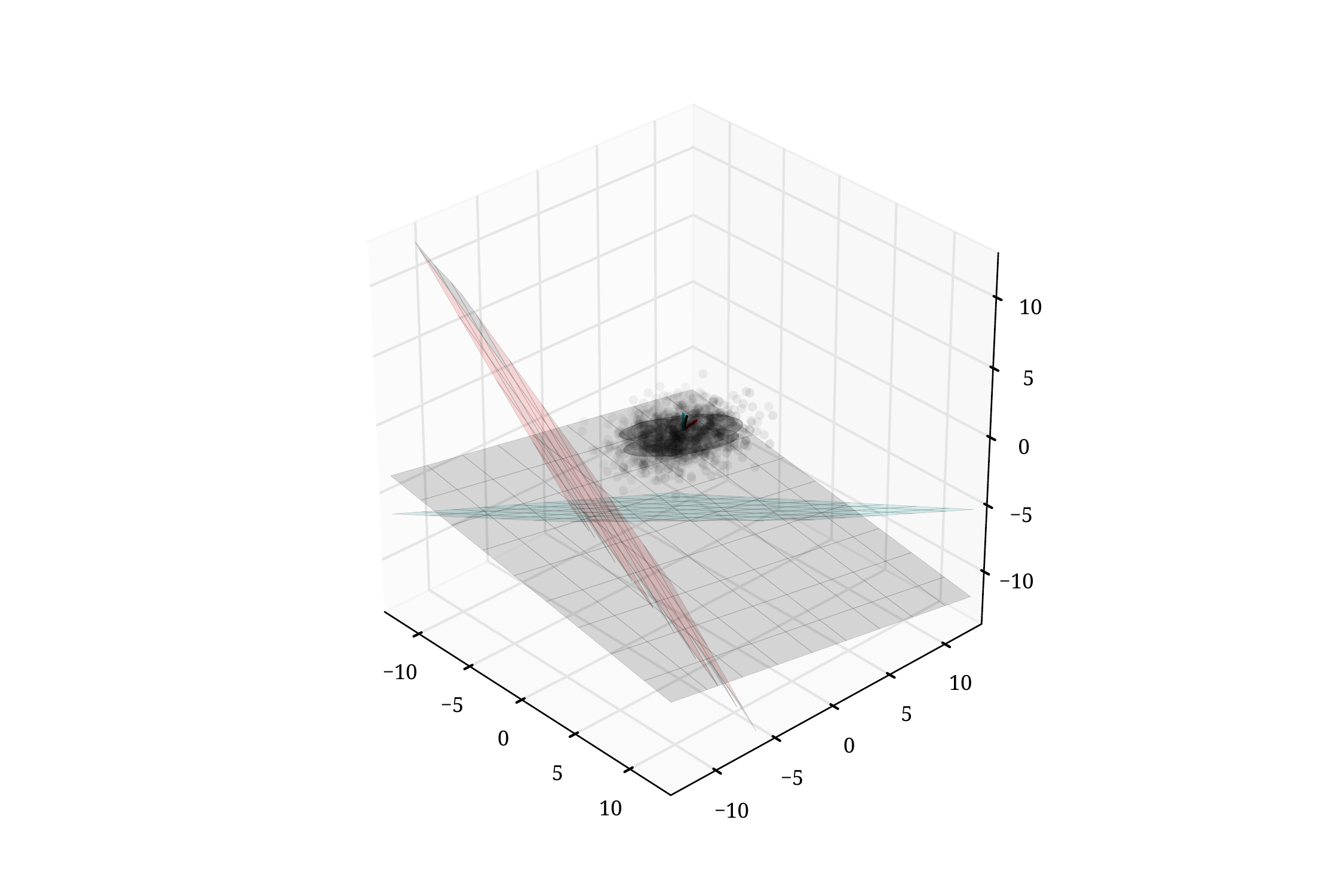}
\end{subfigure}
\begin{subfigure}{0.4\textwidth}
 \centering
 \includegraphics[width=\textwidth,clip=true,trim=20mm 7mm 20mm 0mm]{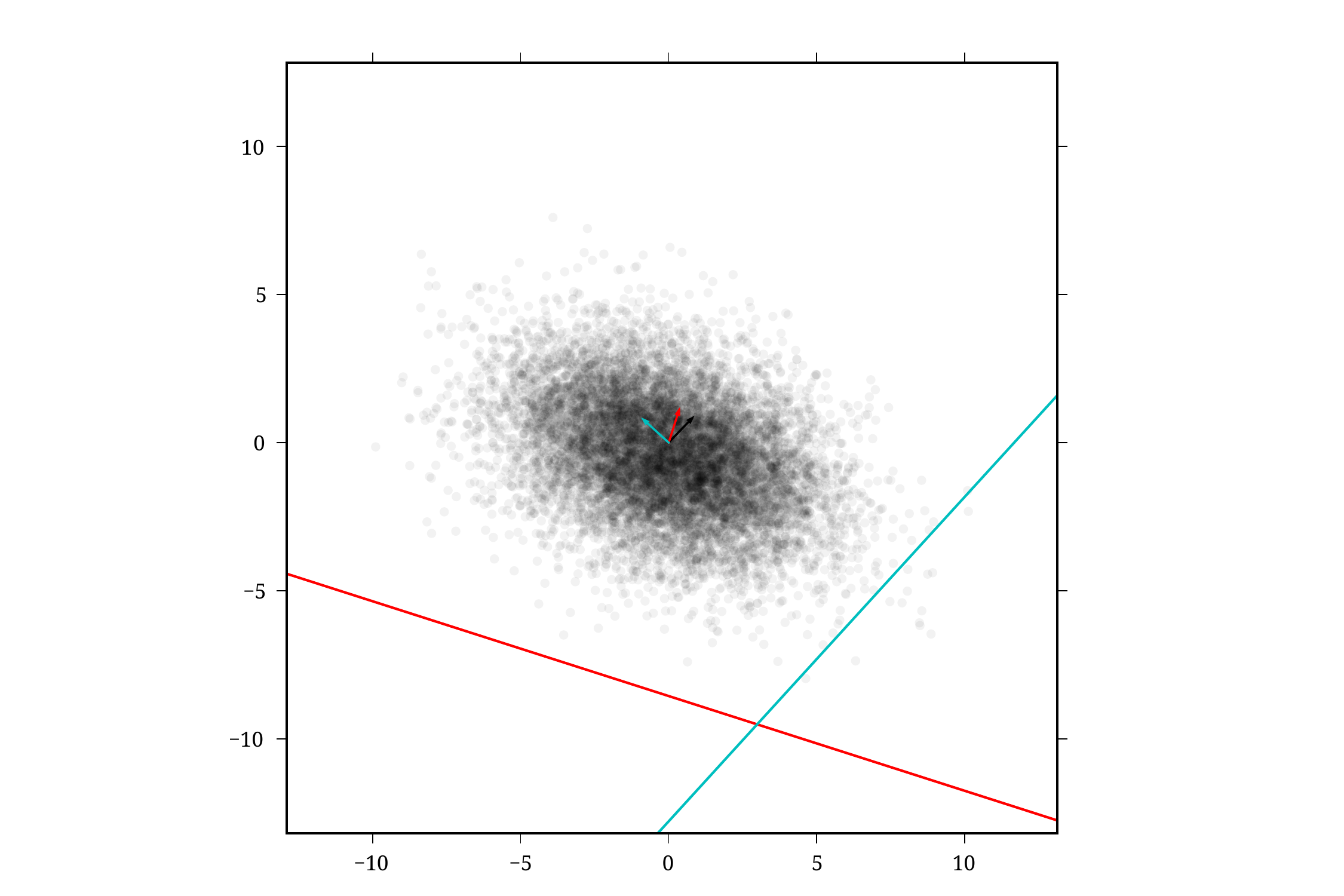}
\end{subfigure}
\caption{Rectified linear feature planes for a 3D gaussian dataset, using a
 complete dictionary ($k=3$). With an undercomplete or $1\times$ complete
 dictionary, these planes tend to move to the bounds of the dataset, forming a
 new coordinate system. The image on the left is a full 3D view of the space,
 while the image on the right is a projection onto the $x$--$y$ plane.}
\label{fig:separators:box}
\end{figure*}

Mirrored by the geometric interpretation, from the perspective of a data point,
nonzero bias terms can be thought of as translating the data to a new origin
along the axis of that feature. Seen from the perspective of a feature, the bias
terms can help move the active region for that feature to an area of space where
data are available to describe---potentially ignoring other regions of space
that contain other portions of the data. From either viewpoint, the bias
parameters in the model can thus help compensate for non-centered data. Figure
\ref{fig:separators:box} shows typical behavior for a $1\times$ complete
dictionary of rectified linear units when trained on an artificial 3D gaussian
dataset. In this case, each feature plane uses its associated bias to translate
away from the origin, together providing a sort of ``bounding box'' for the data.
Though the axes of the box are not necessarily orthogonal, the features define a
new basis that is often aligned with or near the principal axes of the data.

\begin{figure*}[tb]
\centering
\begin{subfigure}{0.5\textwidth}
 \centering
 \includegraphics[width=\textwidth,clip=true,trim=20mm 10mm 20mm 10mm]{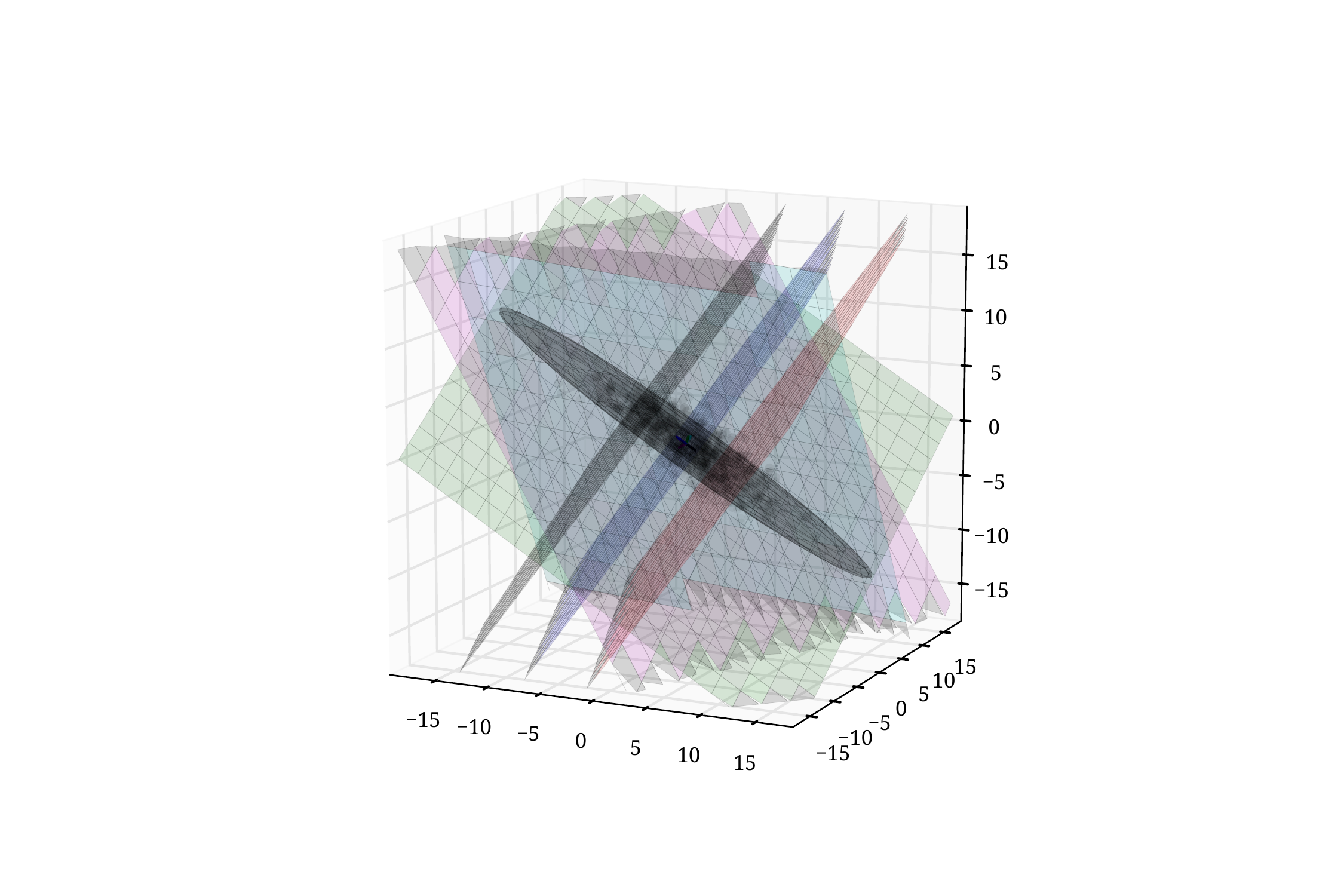}
\end{subfigure}
\begin{subfigure}{0.4\textwidth}
 \centering
 \includegraphics[width=\textwidth,clip=true,trim=20mm 7mm 20mm 0mm]{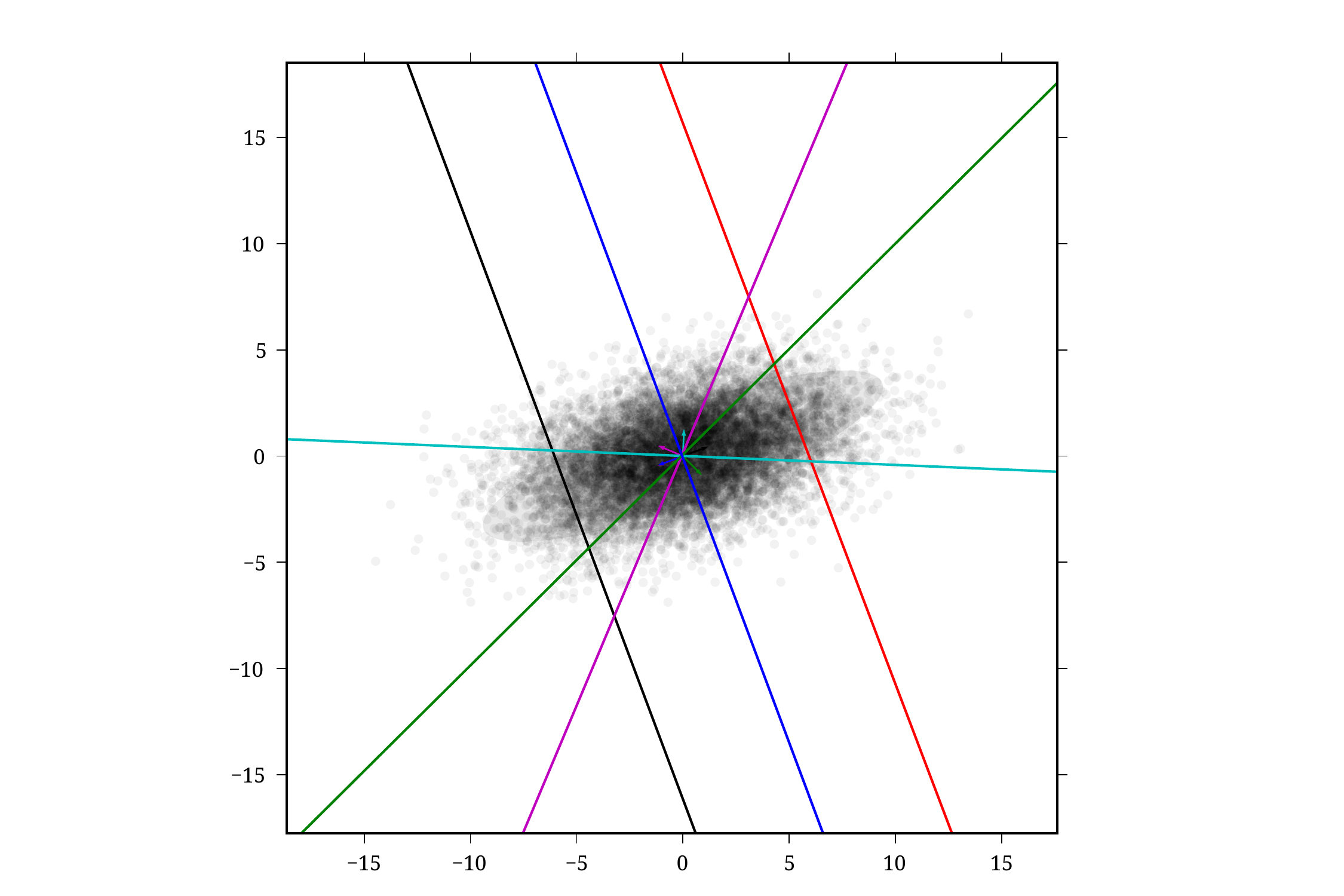}
\end{subfigure}
\caption{Sigmoid feature planes for a 3D gaussian dataset, using an
 overcomplete dictionary ($k=6$). The weights corresponding to each hidden
 neuron form the normal vector to each plane, and the distance from the origin
 is given by the bias for that unit. In the above example, three of the encoding
 planes have partitioned the major axis of the data distribution (perhaps to
 maximize effective coding regions), while the remaining planes capture variance
 along minor axes of the dataset. The image on the left is a full 3D view of the
 space, while the image on the right is a projection onto the $x$--$y$ plane.}
\label{fig:separators:logistic}
\end{figure*}

\begin{figure*}[tb]
\centering
\begin{subfigure}{0.5\textwidth}
 \centering
 \includegraphics[width=\textwidth,clip=true,trim=20mm 10mm 20mm 10mm]{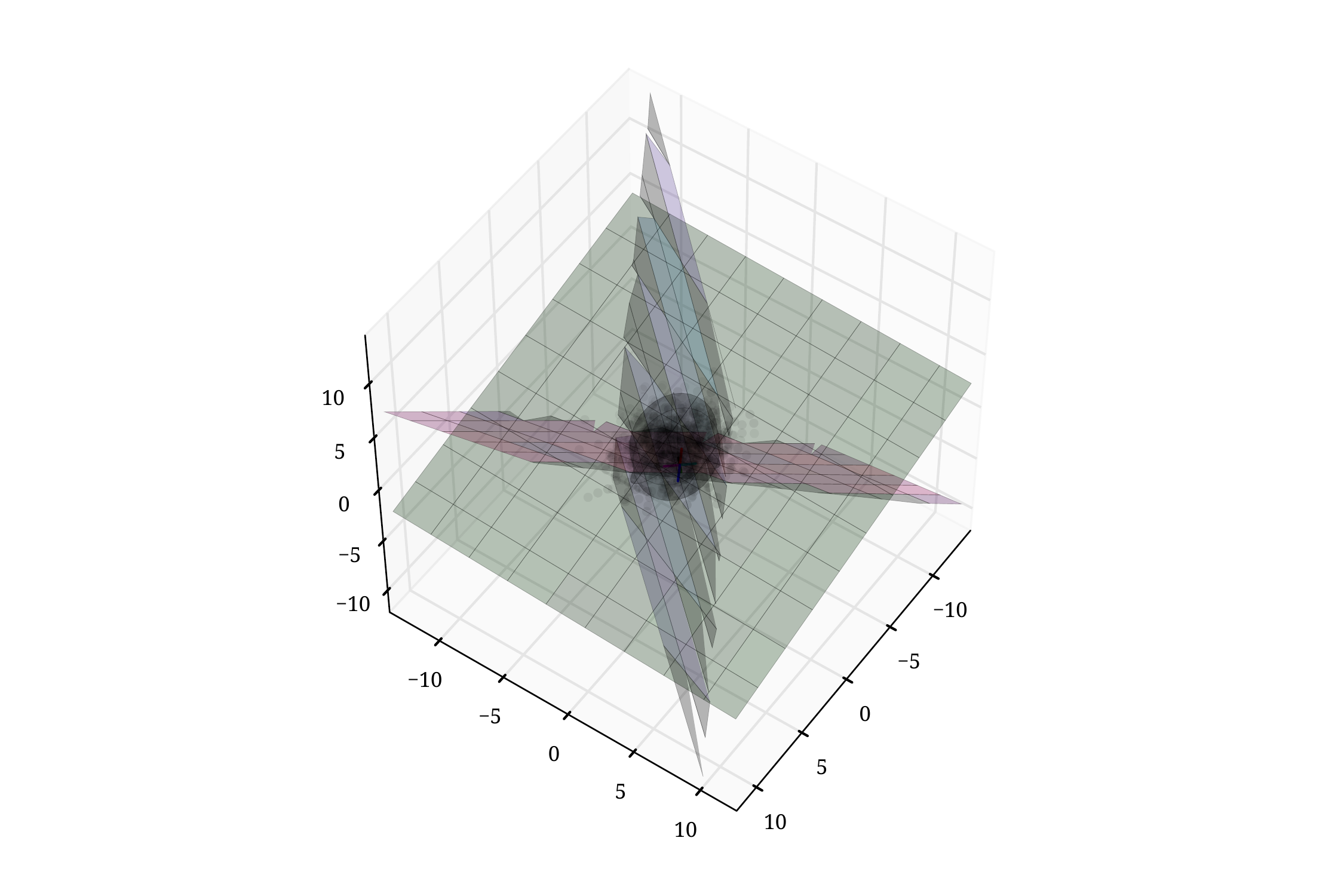}
\end{subfigure}
\begin{subfigure}{0.4\textwidth}
 \centering
 \includegraphics[width=\textwidth,clip=true,trim=20mm 7mm 20mm 0mm]{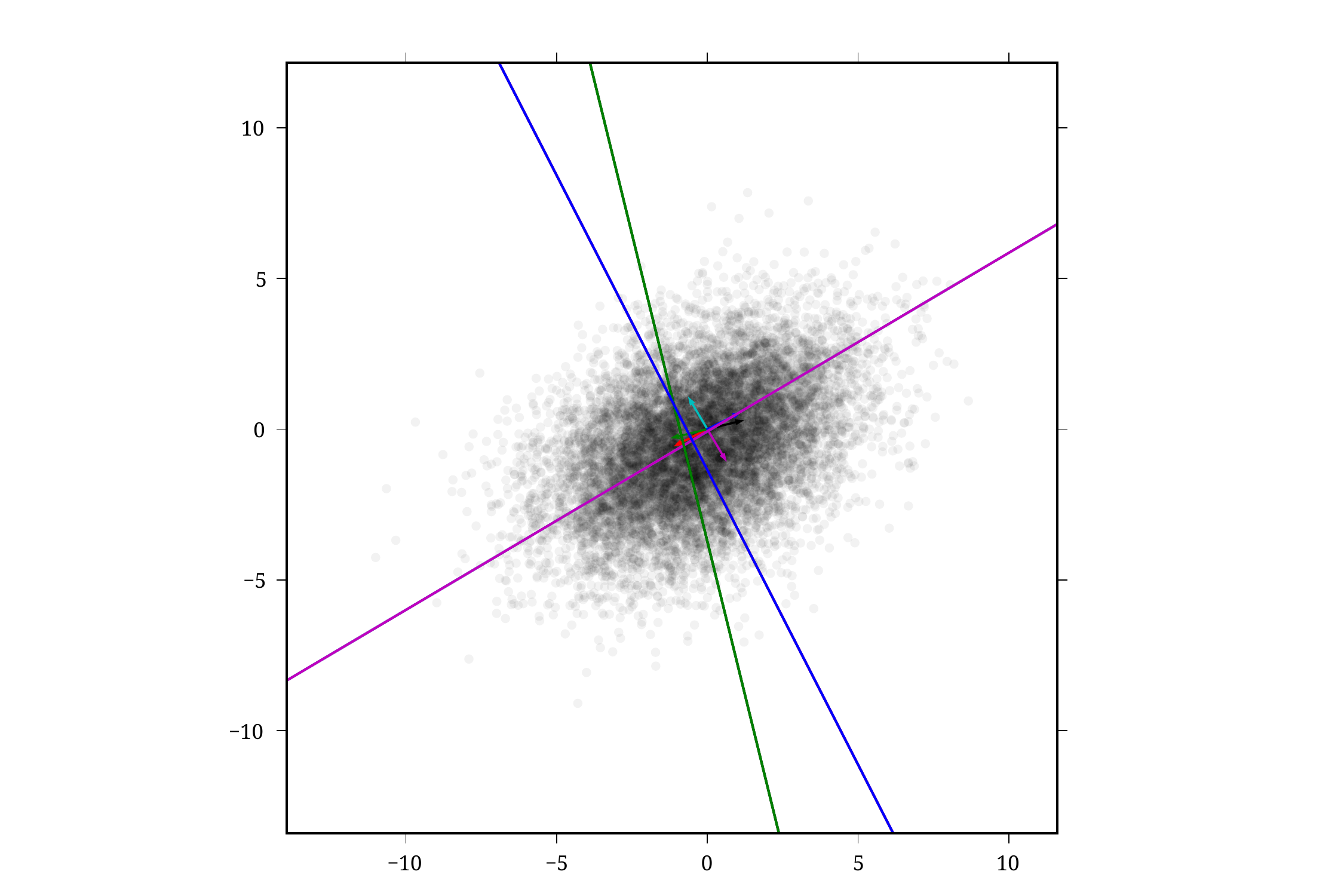}
\end{subfigure}
\caption{Rectified linear feature planes for a 3D gaussian dataset, using an
 overcomplete dictionary ($k=6$). These feature planes group together in pairs,
 splitting the dataset along one axis. The image on the left is a full 3D view
 of the space, while the image on the right is a projection onto the $x$--$y$
 plane.}
\label{fig:separators:split}
\end{figure*}
\subsection{Effective coding region}
\label{sec-3_1}

Visualizing feature hyperplanes is actually instructive for many types of
activation functions. Figure \ref{fig:separators:logistic}, for instance, shows
the feature planes for a $2\times$ overcomplete tied-weights autoencoder with
sigmoid hidden units, trained on an artificial 3D gaussian dataset.\footnote{The
feature planes in this case are aligned with the logistic activation value 0.5.}
Several of the features learned by this model partition the data along the
principal axis of variance, while the remaining features capture some of the
variance along minor axes of the dataset. In comparison, a $2\times$
overcomplete rectified linear autoencoder trained on a 3D gaussian dataset
(Figure \ref{fig:separators:split}) creates pairs of negated feature vectors, in
effect providing a full-wave rectified linear output along each axis. Together,
these pairs of planes split the data into orthants, which produces a coding
scheme that is 50 \% sparse because only half of the features are active for any
given region of the input space.

Clearly these two activation functions have different ways of modeling the input
data, but why do these particular patterns of organization appear? In our view,
the unbounded linear output of the rectified linear autoencoder units addresses
a subtle but important issue inherent in coding with sigmoid activation
functions: scale. Consider, for example, a two-dimensional dataset consisting of
points along a one-dimensional manifold \[ \mathcal{L} = \left\{ [x,
\epsilon]^T : 0 < x < S \right\} \] where $\epsilon$ is gaussian noise.
Regardless of the activation function, this network requires just one hidden
unit to represent points in $\mathcal{L}$: we can set the bias to $0$ and
weights to $[1, 0]^T$ to align the feature of the hidden unit with the linear
function that describes the data. Then points in the dataset are simply coded by
their distance from the origin, as given by the output of the activation
function. For a sigmoid activation function, however, as $S$ increases, the
activation function saturates, limiting the power of the hidden units to
discriminate between small changes in $x$ near $S$. This saturation limits the
scale at which sigmoid units can describe data effectively, but it also prevents
numerical instability by limiting the range of the output.

Regardless of the activation function, then, each feature vector can be seen as
normal to a hyperplane in input space. The activation function encodes data
along the axis of the feature as a function of the distance to the hyperplane
for that feature. Sigmoid activation functions saturate for points far from the
hyperplane, whereas linear activations solve this ``vanishing signal'' problem by
coding all $x$ values with a linear output response. Rectified linear
activations make a further distinction by coding only one half of the input
space. However, the outputs of a linear activation are unbounded, so networks
with inputs at large scales are prone to numerical instability during training.
\section{Behavior on complex data}
\label{sec-4}

\begin{figure*}[tb]
\centering
\begin{subfigure}{0.3\textwidth}
 \centering
 \includegraphics[width=\textwidth,clip=true,trim=0mm 0mm 0mm 0mm]{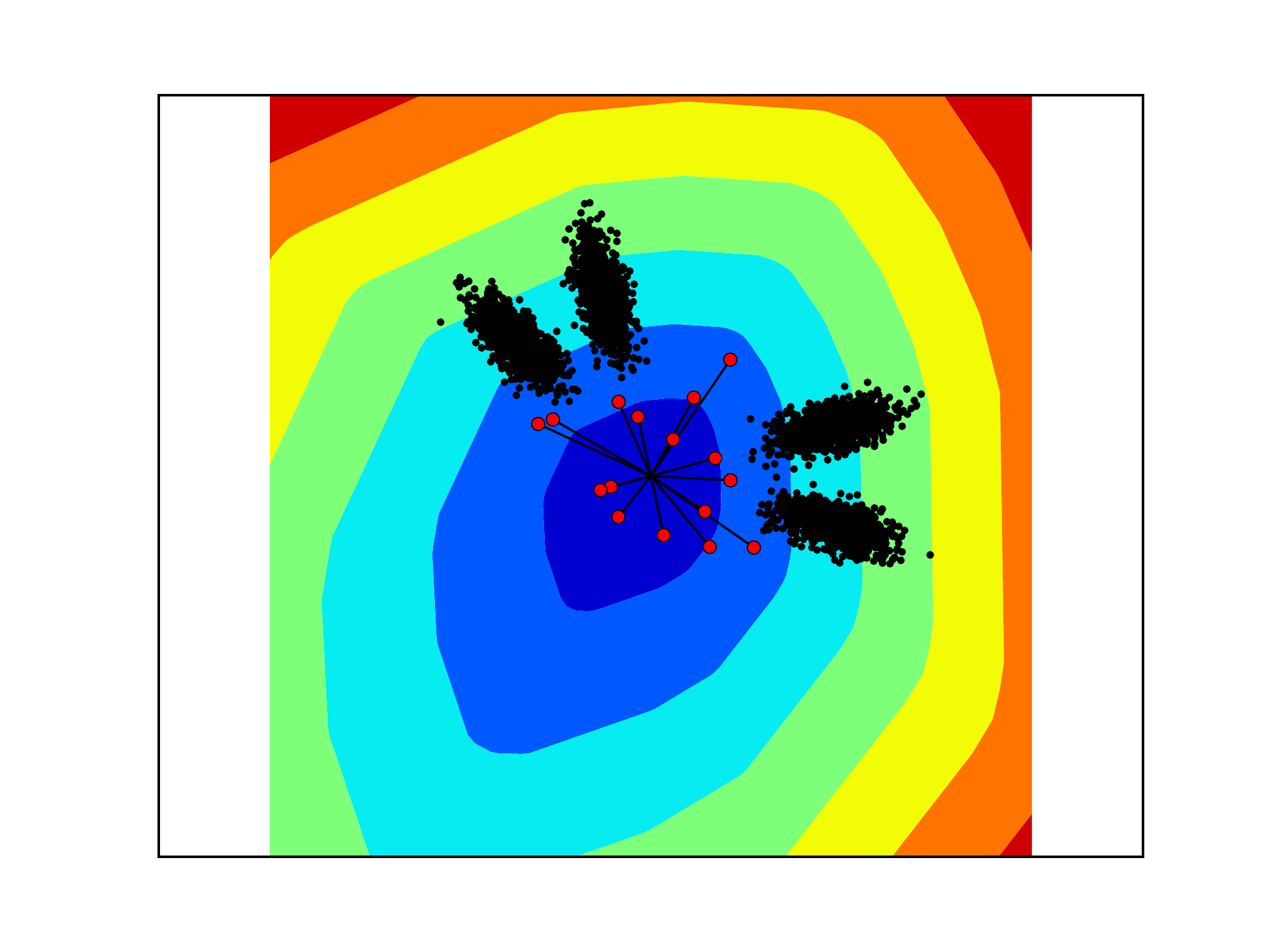}
\end{subfigure}
\begin{subfigure}{0.3\textwidth}
 \centering
 \includegraphics[width=\textwidth,clip=true,trim=0mm 0mm 0mm 0mm]{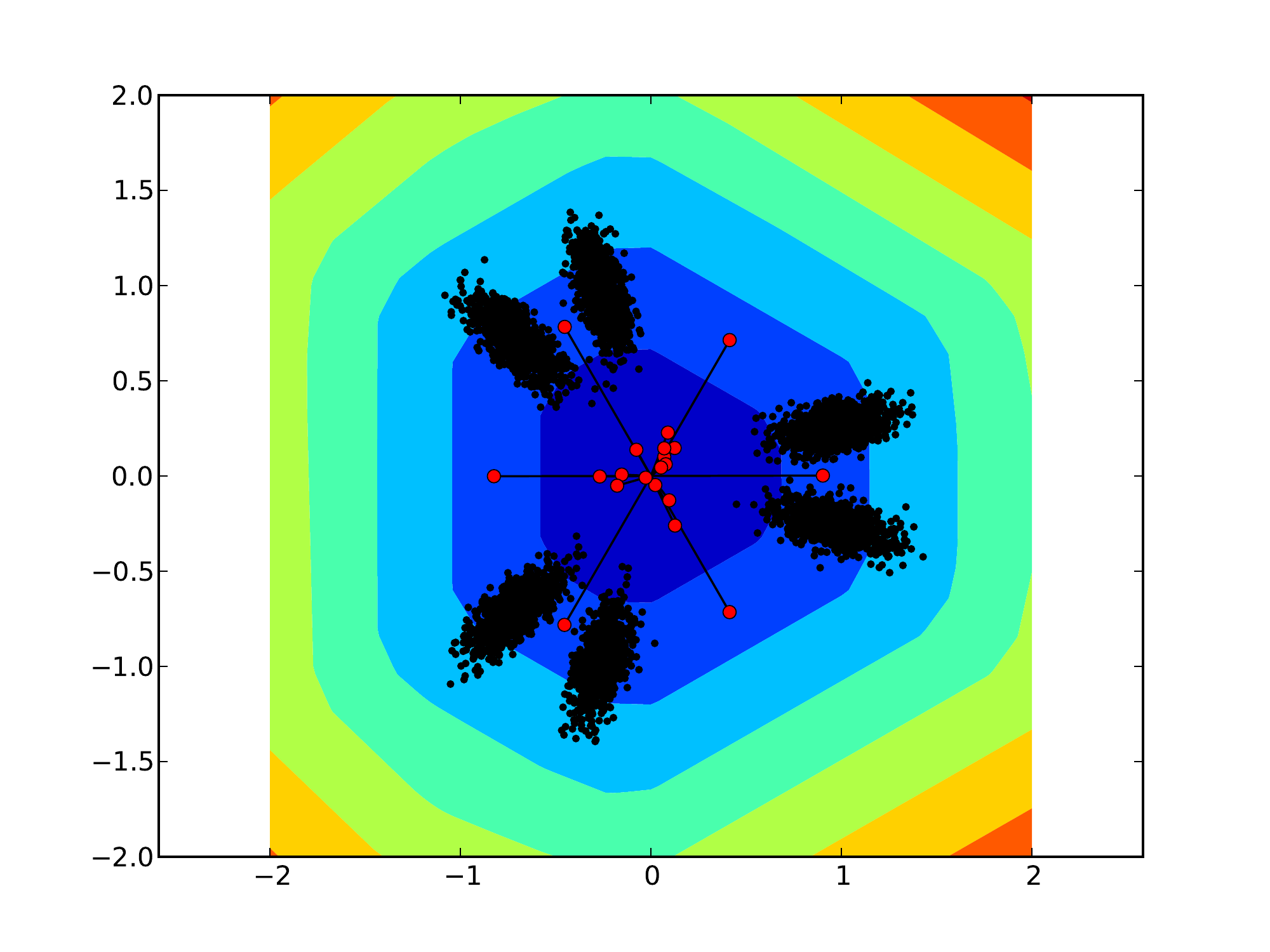}
\end{subfigure}
\begin{subfigure}{0.3\textwidth}
 \centering
 \includegraphics[width=\textwidth,clip=true,trim=0mm 0mm 0mm 0mm]{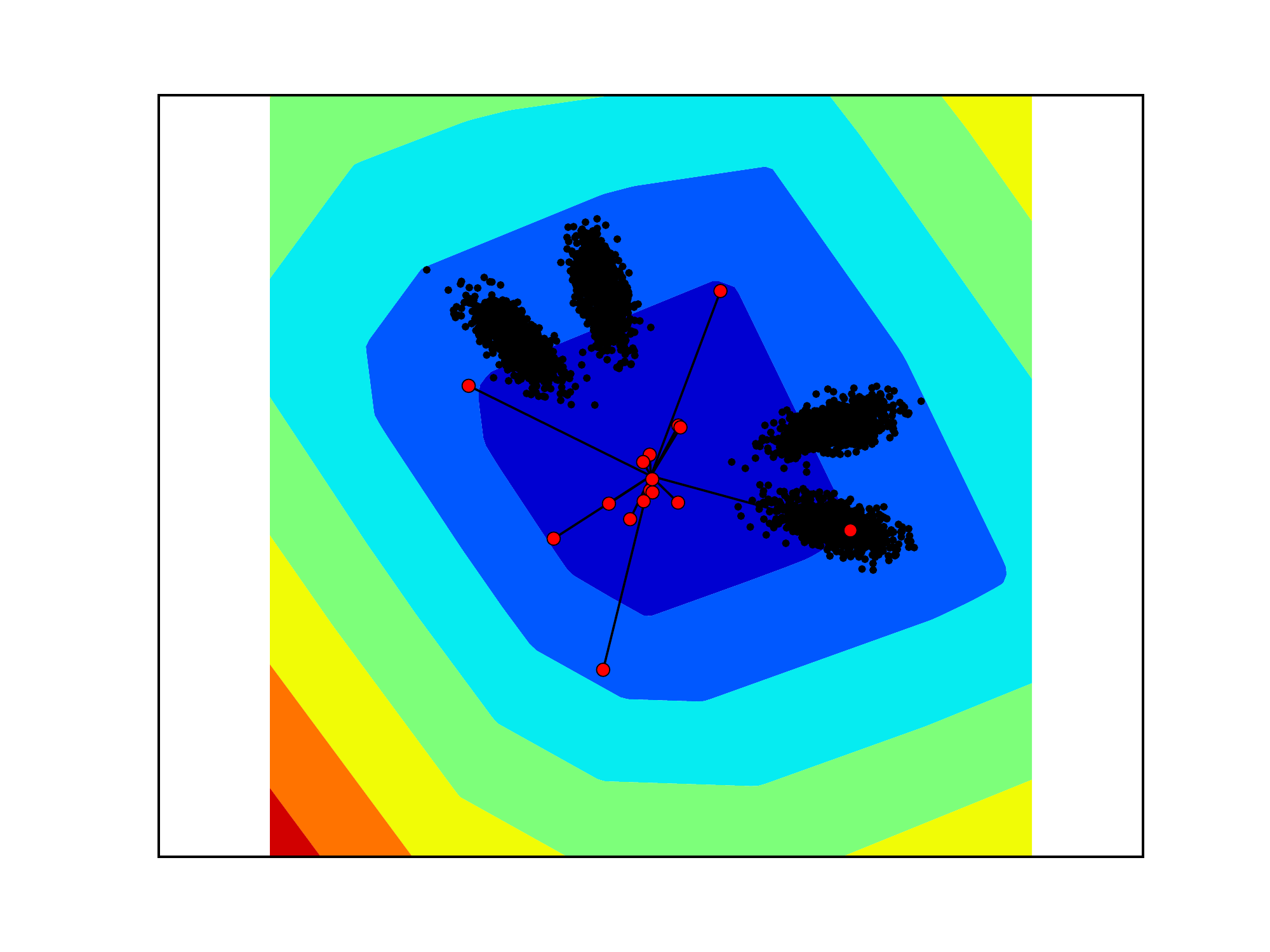}
\end{subfigure}
\caption{Rectified linear responses summed over input space for mixture of
 gaussian datasets. The plot on the left was trained with no regularization,
 while the plots in the middle and on the right were trained with an L1 sparsity
 penalty on the hidden activations. Color delineates regions of equal potential
 summed over all hidden units, and small red circles indicate the learned
 feature vectors. Data points are shown in black.}
\label{fig:mog:contours}
\end{figure*}

\begin{figure*}[tb]
\centering
\includegraphics[width=0.8\textwidth,clip=true,trim=0mm 10mm 0mm 10mm]{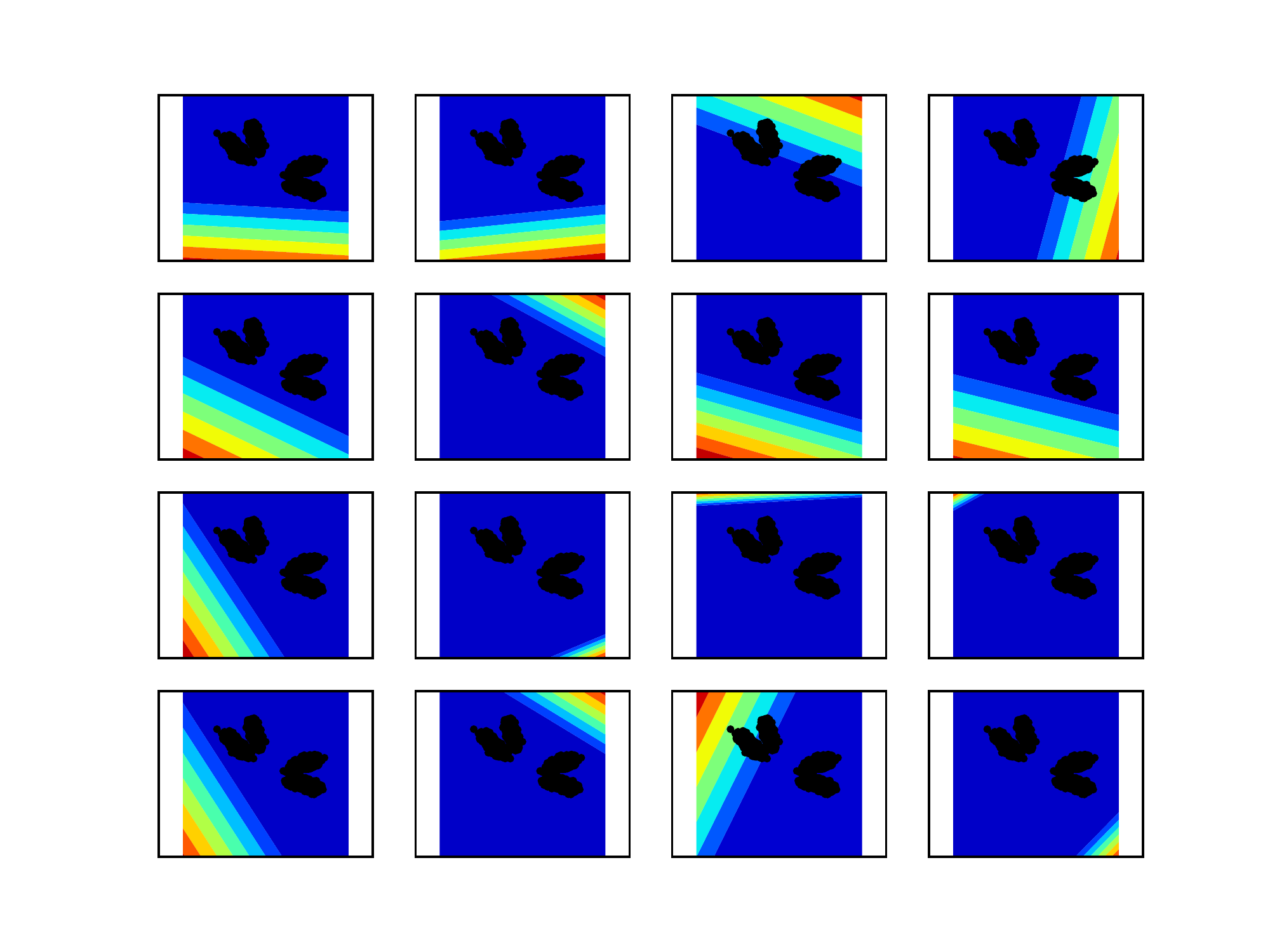}
\caption{Summary plots of individual feature response regions for a rectified
 linear autoencoder trained on a mixture of gaussians dataset, including an L1
 sparsity penalty. Colored bands indicate equipotential regions of response for
 the feature, and data points are shown in black. Interestingly, most of the
 features are inactive for this dataset, with only three features devoted to
 coding important regions of the data.}
\label{fig:mog:features}
\end{figure*}

So far, the visualizations tools that we have used are based on simple gaussian
distributions of data. Many interesting datasets, however, are emphatically not
gaussian---indeed, one of the primary reasons that ICA is so effective is that
it explicitly searches for a non-gaussian model of the data! In this section
we present several visualizations for these models using more complex datasets:
first we examine mixtures of gaussians, and then we use the MNIST digits dataset
as a foray into a more natural set of data.

We first tested the behavior of rectified linear autoencoders on a small, 2D
mixture of gaussians dataset. (See Figure \ref{fig:mog:contours}.) After
training, even massively overcomplete dictionaries tended to display the feature
pairing behavior described above, particularly when combined with a sparse
regularizer, suggesting that these networks are capable of identifying the
number of features required to code the data effectively. Unfortunately, while
it provides simple visualization, working in 2D does not necessarily generalize
to higher dimensions, as our intuition in low-dimensional spaces can quickly
lead us astray in larger spaces.
\subsection{MNIST digits}
\label{sec-4_1}

\begin{figure*}[tb]
\centering
\includegraphics[width=0.5\textwidth,clip=true,trim=0mm 0mm 0mm 0mm]{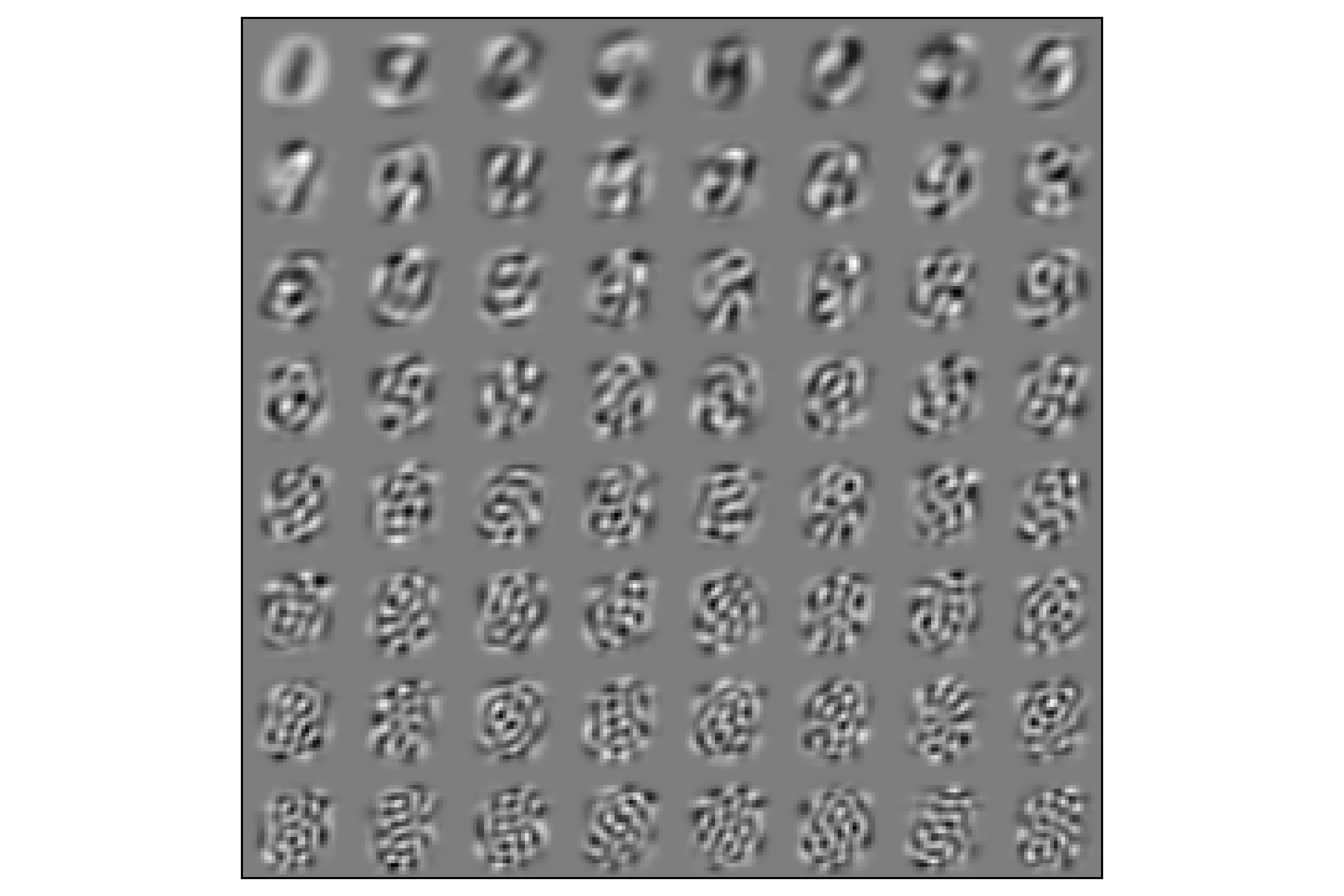}
\caption{PCA features (eigendigits) computed for MNIST digits.}
\label{fig:eigendigits}
\end{figure*}

\begin{figure*}[tb]
\centering
\includegraphics[width=0.8\textwidth,clip=true,trim=0mm 0mm 0mm 0mm]{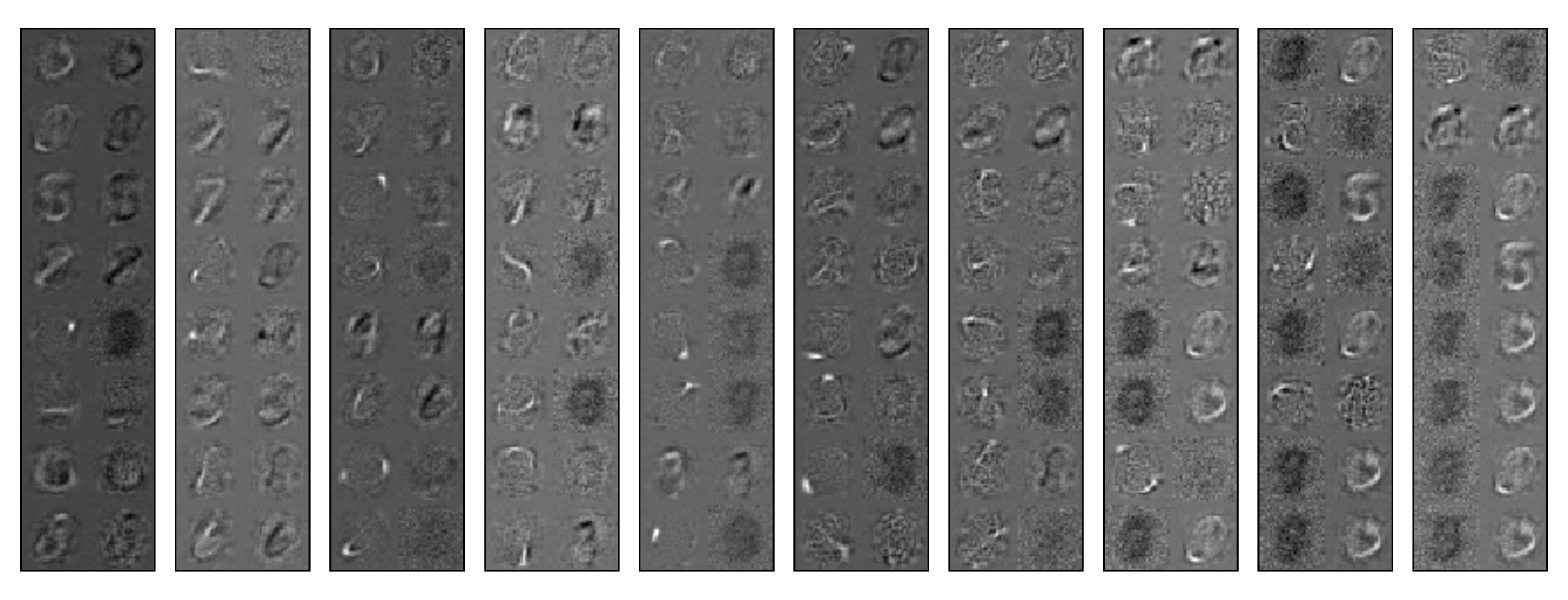}
\caption{Some of the features computed using a single-layer rectified linear
 autoencoder with a $1.5\times$ overcomplete dictionary. Each feature ${\bf w}$
 is displayed on the left, and on the right is its corresponding maximally
 negative feature ${\bf v} = \argmin_{\bf u} {\bf w}^T{\bf u}$.}
\label{fig:relu-digits}
\end{figure*}

\begin{figure*}[tb]
\centering
\begin{subfigure}{0.4\textwidth}
 \centering
 \includegraphics[width=\textwidth,clip=true,trim=0mm 0mm 0mm 0mm]{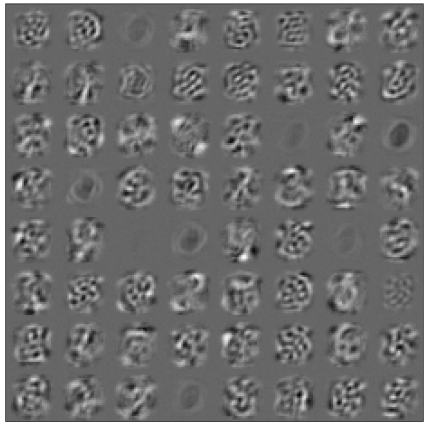}
 \caption{64 first-layer features.}
\end{subfigure}\hspace{2em}
\begin{subfigure}{0.5\textwidth}
 \centering
 \includegraphics[width=\textwidth,clip=true,trim=0mm 0mm 0mm 0mm]{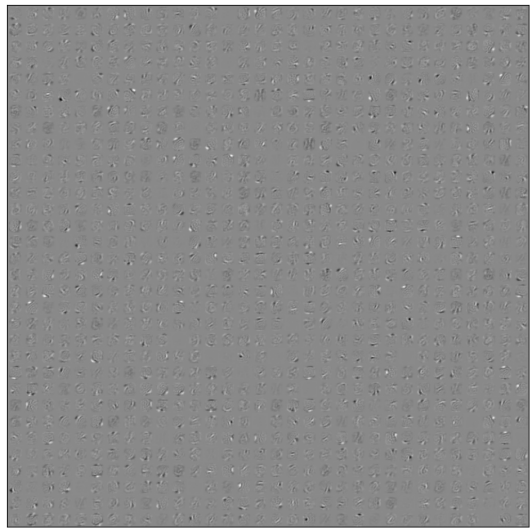}
 \caption{1024 second-layer features.}
\end{subfigure}
\caption{MNIST features computed using a two-layer rectified linear autoencoder.}
\label{fig:two-layer-relu-digits}
\end{figure*}

The MNIST dataset consists of 60000 $28 \times 28$ grayscale images of the
handwritten digits 0 through 9. Figure \ref{fig:eigendigits} shows the PCA
``eigendigits'' computed using this dataset; these digits point in the directions
of highest variance for this dataset overall, and are often interpreted as a
coding of the digit using a Fourier basis. Figure \ref{fig:relu-digits} shows
the features with the highest bias values computed by a single-layer rectified
linear autoencoder on the same dataset. For visualization purposes, each feature
${\bf w}$ on the left of each column is paired with its maximally negative
feature ${\bf v} = \argmin_{\bf u} {\bf w}^T{\bf u}$ from the dictionary. Many
of the primary features in the dictionary are negative images of each other,
indicating that the ``pairing'' of feature planes observed for low-dimensional
datasets also occurs with higher-dimensional data.

Finally, Figure \ref{fig:two-layer-relu-digits} shows features learned by a
two-layer rectified linear autoencoder with untied weights. This model was
trained with just 64 first-layer hidden units, and 1024 second-layer units. As
observed in the low-dimensional gaussian data, the features from the first layer
appear to model the principal axes of the data, providing a bounding box of 64
dimensions that encodes the data for the next layer. The features from the
second layer resemble a more traditional sparse code, consisting in this case of
small segments of pen strokes. It is important to note that the training dataset
was not whitened beforehand, and no regularization was used to train the
network---the autoencoder learned these features automatically, using only a
squared error reconstruction cost.
\section{Conclusion}
\label{sec-5}

This paper has synthesized many recent developments in encoding and neural
networks, with a focus on rectified linear activation functions and whitening.
It also presented an intuitive interpretation for the behavior of these
encodings through simple visualizations. Sample features learned from both
artificial and natural datasets provided examples of learning behavior when
these algorithms are applied to different types of data.

There are many more paths to explore in this area of machine learning research.
Whitening has appeared in many places throughout this paper and seems to be a
very important component of linear autoencoders and coding systems in general.
However, this connection seems poorly understood, and would benefit from further
exploration, particularly with respect to the idea of whitening local regions of
the input space.

\bibliographystyle{plain}
\bibliography{/Users/leif/Papers/annotated}

\end{document}